\documentclass[unnumsec,webpdf,contemporary,large]{oup-authoring-template}%
\usepackage{booktabs}
\usepackage{svg}
\usepackage{paralist}
\usepackage{tabularx}
\usepackage{makecell}
\usepackage{colortbl}
\usepackage{subfig}

\newcolumntype{x}[1]{>{\centering\arraybackslash\hspace{0pt}}p{#1}}
\graphicspath{{Fig/}}

\theoremstyle{thmstyleone}%
\theoremstyle{thmstyletwo}%
\theoremstyle{thmstylethree}%

\begin{document}

\journaltitle{preprint}
\DOI{-}
\copyrightyear{2025}
\pubyear{}
\access{}
\appnotes{Paper}

\firstpage{1}


\title[Knowledge-augmented PLMs for Biomedical RE]{Knowledge-augmented Pre-trained Language Models for Biomedical Relation Extraction}

\author[1,2,$\dagger$]{Mario Sänger\ORCID{0000-0002-2950-2587}}
\author[1,$\ast$]{Ulf Leser\ORCID{0000-0003-2166-9582}}

\authormark{Sänger and Leser}

\address[1]{\orgdiv{Knowledge Management in Bioinformatics}, \orgname{Humboldt-Universität zu Berlin}, \orgaddress{\street{Unter den Linden 6}, \postcode{10099}, \state{Berlin}, \country{Germany}}}
\address[2]{\orgdiv{Enterprise AI Services}, \orgname{AstraZeneca}, \orgaddress{\street{Av. Diagonal 615}, \postcode{08028}, \state{Barcelona}, \country{Spain}}}

\corresp[$\ast$]{Corresponding author: \{leser\}@informatik.hu-berlin.de  

$\dagger$ The study was conducted entirely while M.S. was employed at Humboldt-Universität zu Berlin.
}

\abstract{%
Automatic relationship extraction (RE) from biomedical literature is critical for managing the vast amount of scientific knowledge produced each year. 
In recent years, utilizing pre-trained language models (PLMs) has become the prevalent approach in RE. 
Several studies report improved performance when incorporating additional context information while fine-tuning PLMs for RE.
However, variations in the PLMs applied, the databases used for augmentation, hyper-parameter optimization, and evaluation methods complicate direct comparisons between studies and raise questions about the generalizability of these findings.
Our study addresses this research gap by evaluating PLMs enhanced with contextual information on five datasets spanning four relation scenarios within a consistent evaluation framework. 
We evaluate three baseline PLMs and first conduct extensive hyperparameter optimization. 
After selecting the top-performing model, we enhance it with additional data, including textual entity descriptions, relational information from knowledge graphs, and molecular structure encodings.
Our findings illustrate the importance of i) the choice of the underlying language model and ii) a comprehensive hyperparameter optimization for achieving strong extraction performance. 
Although inclusion of context information yield only minor overall improvements, an ablation study reveals substantial benefits for smaller PLMs when such external data was included during fine-tuning. }
\keywords{Biomedical Natural Language Processing, Relation extraction, Pre-trained Language Models, Knowledge Augmentation, Benchmark}

\maketitle

\section{Introduction}
Automatic relationship extraction (RE) from biomedical literature is an essential tool for managing and synthesizing the ever-increasing volume of scientific knowledge produced annually. 
This technology enables researchers to identify and analyze large sets of  complex interactions between genes, diseases, drugs, and other biomedical entities within the literature, significantly accelerating the pace of scientific discovery \citep{wei2024pubtator}.
For instance, in fields such as pharmacovigilance and drug development, accurately identifying potential drug interactions is vital as these interactions can lead to adverse effects or diminished therapeutic efficacy \citep{harpaz2014text}. 
Accordingly, methods for extracting relationships from biomedical texts have been investigated intensively over the last two decades \citep{zhou2014biomedical}.

In recent years, pre-trained language models (PLMs) have become the prevalent technology in relation extraction \citep{weber2022chemical,lai2023biorex}.
These methods usually leverage a transfer learning setting, i.e., the language model is pre-trained on extensive in-domain text collections first and then fine-tuned to the specific relation extraction task using human-labeled gold standard data sets \citep{lee2020biobert}.
For improving the fine-tuning process, several studies explore utilizing additional context information,
including curated entity definitions \citep{asada2021representing}, knowledge graph triplets \citep{sousa2023k,asada2023integrating}, and chemical structure information \citep{asada2021using}, for improving the fine-tuning of PLM-based relation extraction methods.
For example,  Aldahdooh et al. \citep{aldahdooh2024mining} combine PLMs with gene descriptions from the Entrez Gene database \citep{maglott2010entrez} and chemical descriptions from the Comparative Toxicogenomics Database (CTD) \citep{ComparativeToxDavis2023} for improved mining drug-target interactions.
Dou et al. \citep{dou2023ik} utilize human-curated descriptions of different drug aspects from DrugBank to enhance a PLM-based baseline model for drug-drug interaction prediction.
Furthermore, Asada et al. \cite{asada2023integrating} leverage knowledge graph information to augment a PubMedBERT-based drug-drug interaction identification model.
Similarly, Sousa et al. \cite{sousa2023k} explore utilizing class and relational data from four different knowledge bases for drug-drug, chemical-disease, and phenotype-gene relation extraction using SciBERT~\citep{beltagy2019scibert}.
Most of these studies report strong advantages when incorporating additional contextual information when compared to approaches that train the PLMs exclusively on annotated texts \citep{aldahdooh2024mining,dou2023ik,sousa2023k,asada2023integrating,mcinnes2024biobert}.
A direct comparison of published results, however, usually is not possible due to technical differences (e.g., choice of the underlying language model, used databases, model design) and applied evaluation procedure.
This raises questions about the determining factors for the improvements achieved and the transferability of the studies' insights to other RE scenarios and to other PLMs.
For instance, our investigation in \citep{weber2022chemical} explored knowledge-enhanced PLMs for identifying chemical-protein relations and found that context information achieves only marginal improvements if the hyperparameters of the PLM were carefully optimized.
It thus  an open research question whether RE based on PLMs actually benefit from additional information or not - and under which circumstances.

In this study we address this research gap by assessing PLMs augmented with different types of additional context information in different relation extraction use cases.
We utilize five data sets encompassing four biomedical relation scenarios within a uniform evaluation setting, allowing to derive more reliable insights into the effectiveness of including context information in the fine-tuning process of PLMs.
For each data set, we perform an extensive hyperparameter optimization of three baseline PLM models (i.e., PubMedBERT~\citep{gu2021domain}, RoBERTa-large-PM-M3-Voc~\citep{lewis2020pretrained}, BioLinkBERT-Large~\citep{yasunaga2022linkbert})  first, before extending the best model with additional data.
We explore the inclusion of (a) textual entity descriptions, (b) embedded information representing an entity's neighborhood in a knowledge graph as well as its mentions in the literature \cite{sanger2021large}, and (c) molecular structure encodings for drug- and chemical-related scenarios.

Our experimental results highlight a superior performance of the BioLinkBERT-large model across all extraction scenarios. 
However, we mostly achieve no or only minor performance improvements when including additional context information in the fine-tuning procedure compared to adapting the model on gold standard annotations only.
To verify this result, that potentially contradicts several previous studies, we perform an ablation study on the sizes of the PLM models used, and reveal that the extraction quality of PLM-based models having fewer parameters benefit considerably from incorporating additional external information while fine-tuning - an effect that vanishes with more recent and considerably larger models. 
These findings suggest that the larger PLMs implicitly encode (to some extent) the supervision signals from the additional information. 

\section{Material and Method}
\subsection{Data Sets}
\label{5:sec:data-sets}
We investigate our method on five data sets spanning four relationship scenarios, i.e. chemical-disease,  chemical-gene, drug-drug, and gene-disease, providing a more comprehensive evaluation than in prior art. 
We leverage the following datasets:
\begin{itemize}
    \item Chemical-disease: BioCreative-V-CDR (BC5CDR) \citep{li2016biocreative}
    \item Chemical-gene: ChemProt \citep{krallinger2017overview}, CPI~\citep{doring2020automated}
    \item Drug-drug: DDI corpus \citep{herrero2013ddi}
    \item Gene-disease: ChemDisGene \citep{zhang-etal:2022:LREC}

\end{itemize}
Note, we utilize two data sets for chemical-protein relations to analyze and account for possibly existing data set-specific patterns or biases. Two of the five data sets (BC5CDR, ChemDisGene)) have only document-level annotations of relationships, while the other three come with mention annotation. 
See Table~\ref{5:tab:datasets} for basic statistics of the used data sets and refer to Appendix \ref{sec:app:data-sets} for an in-detail description.
In order to link the entity mentions to information in knowledge bases, we normalize the entity mentions to shared ontologies, i.e., NCBI Gene \citep{brown2015gene} for genes, CTD Diseases\footnote{CTD Diseases is also known as MEDIC and represents a subset of MeSH~\citep{lipscomb2000medical} and OMIM~\citep{hamosh2005online}.} \citep{ComparativeToxDavis2023} for diseases, and CTD Chemicals\footnote{CTD chemicals represents a subset of MESH~\citep{lipscomb2000medical}.} \citep{ComparativeToxDavis2023} for chemicals.
For BC5CDR and ChemDisGene, we use the provided gold standard MESH and NCBI Gene annotations.
We apply different strategies for the remaining data sets:
\begin{itemize}
    \item ChemProt: For ChemProt, we utilize the annotations provided by PubTator Central \citep{wei2019pubtator} using the PubMed identifiers of the data set.
    \item CPI: For normalizing protein annotations to NCBI Gene, we leveraged the UniProt identifiers    
        \citep{uniprot2023uniprot} given in the data set and the id-mapping service of the platform\footnote{\url{https://www.uniprot.org/id-mapping}}. 
        Chemical mentions are mapped to MESH by applying a cross-reference table\footnote{\url{https://ftp.ncbi.nih.gov/pubchem/Compound/Extras/CID-MeSH}} on the given PubChem identifiers~\citep{kim2019pubchem} of the data set.
    \item DDI: Since the DDI texts essentially originate from Drugbank, which strongly follows the standard nomenclature, we rely on string matching for mapping the drug mentions to CTD chemicals.
\end{itemize}
See Table~\ref{sec:app:entity-normalization} for statistics on the number of normalized mentions and unique entities.
%
\begin{table*}[tbhp]
  \centering
    \begin{tabular}{rcrr|rrr|rr|rr}
    \toprule
     &
       &
       &
       &
      \multicolumn{3}{c|}{\textbf{Splits}} &
      \multicolumn{2}{c|}{\textbf{Entities}} &
      \multicolumn{2}{c}{\textbf{Relations}}
      \\
    \multicolumn{1}{c}{\textbf{Data Set}} &
      \textbf{Text Type} &
      \multicolumn{1}{c}{\textbf{Instances}} &
      \multicolumn{1}{c|}{\textbf{Tokens}} &
      \multicolumn{1}{c}{\textbf{Train}} &
      \multicolumn{1}{c}{\textbf{Val.}} &
      \multicolumn{1}{c|}{\textbf{Test}} &
      \multicolumn{1}{c}{\textbf{Type}} &
      \multicolumn{1}{c|}{\textbf{Count}} &
      \multicolumn{1}{c}{\textbf{Type}} &
      \multicolumn{1}{c}{\textbf{Count}}
      \\
    \midrule
    \multicolumn{1}{l}{BC5CDR} &
      Abstracts &
      \multicolumn{1}{c}{1.500} &
      \multicolumn{1}{c|}{281.792} &
      \multicolumn{1}{c}{500} &
      \multicolumn{1}{c}{500} &
      \multicolumn{1}{c|}{500} &
      Chemical &
      15.953 &
      Chemical-Disease &
      3.169
      \\
     \multicolumn{1}{l}{(DL)}&
       &
       &
       &
       &
       &
       &
      Disease &
      13.318 &
       &
      
      \\
    \midrule
    \multicolumn{1}{l}{Chemprot} &
      Abstracts &
      \multicolumn{1}{c}{2.482} &
      \multicolumn{1}{c|}{563.091} &
      \multicolumn{1}{c}{1020} &
      \multicolumn{1}{c}{612} &
      \multicolumn{1}{c|}{800} &
      Chemical &
      32.514 &
      Upregulator &
      2.061
      \\
     \multicolumn{1}{l}{(ML)}&
       &
       &
       &
       &
       &
       &
      Gene &
      30.922 &
      Downregulator &
      5.098
      \\
     &
       &
       &
       &
       &
       &
       &
       &
       &
      Agonist &
      497
      \\
     &
       &
       &
       &
       &
       &
       &
       &
       &
      Antagonist &
      741
      \\
     &
       &
       &
       &
       &
       &
       &
       &
       &
      Substrate &
      1.873
      \\
        \midrule
    \multicolumn{1}{l}{CPI} &
      Abstracts$\dagger$ &
      \multicolumn{1}{c}{1.808} &
      \multicolumn{1}{c|}{67.163} &
      \multicolumn{1}{c}{1808} &
      \multicolumn{1}{c}{0} &
      \multicolumn{1}{c|}{0} &
      Chemical &
      3.563 &
      Chemical-Protein &
      2.931
      \\
     \multicolumn{1}{l}{(ML)}&
       &
       &
       &
      \multicolumn{1}{c}{\textit{(1208)}} &
      \multicolumn{1}{c}{\textit{(300)}} &
      \multicolumn{1}{c|}{\textit{(300)}} &
      Gene &
      4.256 &
       &
      
      \\
    \midrule
    \multicolumn{1}{l}{ChemDisGene} &
      Abstracts &
      \multicolumn{1}{c}{523} &
      \multicolumn{1}{c|}{123.767} &
      \multicolumn{1}{c}{523} &
      \multicolumn{1}{c}{0} &
      \multicolumn{1}{c|}{0} &
      Chemical &
      5.739 &
      Marker/Mechanism &
      494
      \\
     \multicolumn{1}{l}{(DL)} &
       &
       &
       &
      \multicolumn{1}{c}{\textit{(300)}} &
      \multicolumn{1}{c}{\textit{(80)}} &
      \multicolumn{1}{c|}{\textit{(123)}} &
      Disease &
      2.931 &
      Therapeutic &
      82
      \\
     &
       &
       &
       &
       &
       &
       &
      Gene &
      5.578 &
       &
      
      \\
    \midrule
    \multicolumn{1}{l}{DDI corpus} &
      Abstracts + &
      \multicolumn{1}{c}{1.017} &
      \multicolumn{1}{c|}{167.148} &
      \multicolumn{1}{c}{651} &
      \multicolumn{1}{c}{0} &
      \multicolumn{1}{c|}{191} &
      Brand &
      1.865 &
      Advise &
      1.047
      \\
     \multicolumn{1}{l}{(ML)}&
      DrugBank &
       &
       &
      \multicolumn{1}{c}{\textit{(550)}} &
      \multicolumn{1}{c}{\textit{(101)}} &
       &
      Drug &
      12.405 &
      Effect &
      2.047
      \\
     &
      Texts &
       &
       &
       &
       &
       &
      Group &
      4.221 &
      Interaction &
      285
      \\
     &
       &
       &
       &
       &
       &
       &
       &
       &
      Mechanism &
      1.621
      \\
    \bottomrule
    \end{tabular}%
    \caption{Overview of basic statistics, entity types and relation types of the data sets used in our evaluation.
    For each data set we report the number of total entity mentions and unique entities, identified by the given entity normalization identifiers of the data set (BC5CDR and ChemDisGene) or unique surface forms (Chemprot, CPI and DDI).
    Moreover, we illustrate whether the corpus provides mention- (ML) or document-level (DL) relation annotations.
    If a data set, does not provide a validation and/or test split, we give the size of the created split in parenthesis below.
    $\dagger$ CPI contains only selected sentences from the abstract and not the entire text.
    }
    \label{5:tab:datasets}
\end{table*}

\subsection{Base Model}
\begin{figure*}
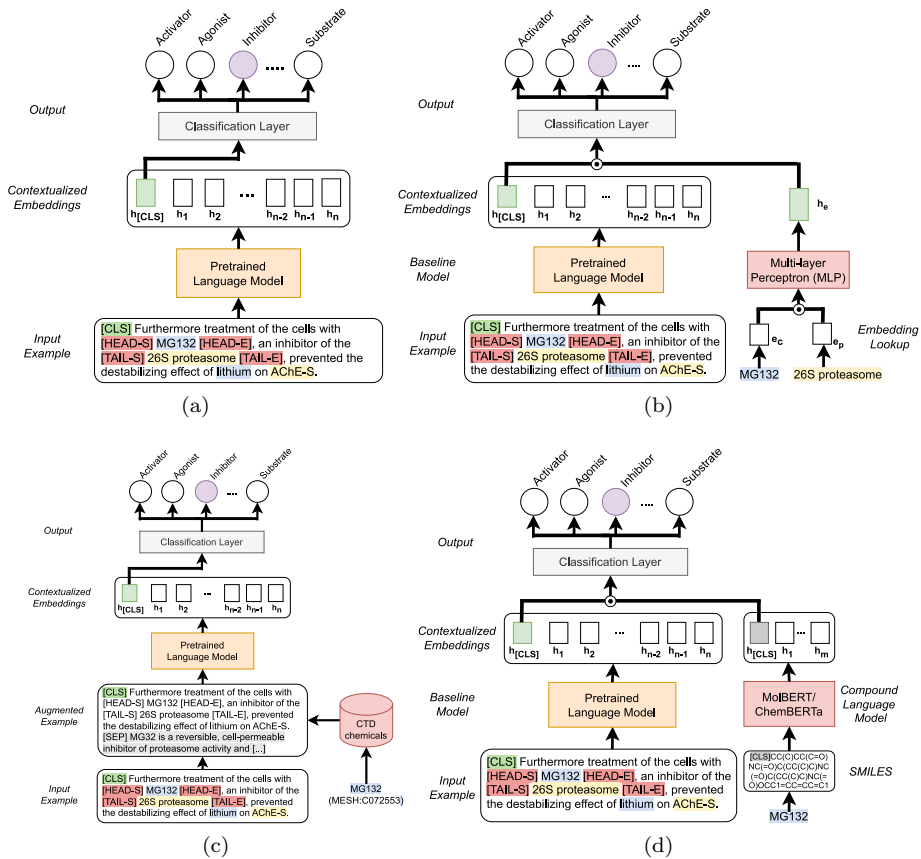

    \centering
    \subfloat[][]{
        \includesvg[height=0.29\textwidth]{figures/model_baseline.svg}
        \label{fig:model_baseline}}
    \subfloat[][]{
        \includesvg[height=0.29\textwidth]{figures/model_kge.svg}
        \label{fig:model_text}}
        
    \subfloat[][]{
        \includesvg[height=0.29\textwidth]{figures/model_text.svg}
        \label{fig:model_kge}}
    \subfloat[][]{
        \includesvg[height=0.29\textwidth]{figures/model_str.svg}
        \label{fig:model_str}}
    \caption{Overview of the baseline model and all extensions we evaluate in our study. 
    (a) Baseline model: 
    We build one input example per entity pair, i.e., chemical-protein pair, in each sentence and mark the pair under investigation with special tokens. 
    The sentence is embedded using a pretrained language model and the [CLS] token embedding is passed through an output layer performing the relation classification. 
    (b) Model enhanced with additional embedded entity information:
    First, we lookup the pre-trained KB embeddings for the head and tail entity under investigation and input the concatenation of both to a multi-layer perceptron (MLP).
    The resulting embedding is concatenated to the contextualized input text embedding as additional input to the classification layer.
    (c) Model augmented with additional textual information: 
    We conduct KB lookups to retrieve textual descriptions for the head entity, the tail entity or both using entity identifiers.
    We include the textual descriptions by appending them to the input text separated by the [SEP] token.
    (d) Model extended with molecular information of chemicals:
    For a given chemical $c$, we first retrieve its respective SMILES string $SM_c$, prepend the [CLS] token to $SM_c$ and feed it into the compound language model.
    We use the representation of the [CLS] token in the last hidden layer as encoding for the structure of $c$ and concatenate it with the language model's output to form the classification layer's input.
}
    \label{fig:re-model}
\end{figure*}

We follow common practice \cite{lee2020biobert} and model relation extraction as a multilabel, sentence-level relation classification problem, which we approach by fine-tuning a pre-trained transformer-based language model. 
More specifically, we generate one training/testing example per pair of entities that occur together in the same sentence. 
We mark the entity pair under investigation by inserting the special tokens [HEAD-S], [HEAD-E], [TAIL-S], and [TAIL-E], which highlight the beginning and end of the head and of the tail entity.
Moreover, we prepend the [CLS] token, a specially designed token to aggregate the information of the entire input text in PLMs, to each input example.
We only form entity pairs that comply with the entity types of the respective relation type. 
We follow the data set standard for DDI by creating only one input instance per drug-drug pair based on the order of occurrence in the input text. 
We use a pre-trained language model to obtain a contextualized embedding $h_i$ of every token in the sentence and represent the sentence by using the embedding of the [CLS] token.
Finally, we apply a linear layer to the sentence representation and transform the activation score with a sigmoid nonlinearity. 
See Figure~\ref{fig:model_baseline} for an illustration of the baseline model.

We inspect two approaches for enhancing the input text separately.
First, we add the sentence before and after the sentence under examination to the input text. 
This augmentation is intended to provide the model with additional textual context to make more informed relation decisions.
Second, we prepend a verbal task instruction to the input text:
%
\begin{quote}
    \textit{Is there a $<$relation-type$>$ interaction between $<$head-entity$>$ and $<$tail-entity$>$?}
\end{quote}
In the instruction, we replace the placeholders $<$relation-type$>$, $<$head-entity$>$, and $<$tail-entity$>$ with the focused relation type and the specific entity mentions of the input example.

\subsection{Model Extensions}
In the following paragraphs, we describe how the base model is augmented with additional entity information aiming to provide contextual knowledge beyond the text under consideration.
We believe that the extended context information can offer valuable insights to the model when deciding upon relationships between entities.
Depending on the type of the entities investigated different information are added, i.e., textual descriptions and embedded information for all considered entity types as well as molecular structure information for drug-/chemical-related use cases.
The inclusion of all three types of additional entity information have been investigated in prior work, e.g., refer to  \citep{aldahdooh2024mining,dou2023ik} for textual, \citep{asada2021representing,sousa2023k} and \citep{tang2024mollm,
luo2024learning} for molecular information.
Yet, there has been no systematic comparison that evaluates all these types within a single, unified setting.

\subsubsection{Additional Textual Information}
\label{5:sec:text_info}
We hypothesize that augmenting the input with additional textual information beyond the sentence context could result in a more accurate model.
For instance, in case of chemical-protein relation extraction, this could include information indicating that a chemical is known to function as an antagonist for a particular group of proteins or that a protein is associated with a specific protein family.

We experiment with diverse additional textual entity information gathered from different database bases, i.e., \textit{CTD chemicals} for chemicals/drugs, \textit{CTD diseases} for disease, and \textit{NCBI gene} for genes. 
%
%
For instance, in cases where the chemical \textit{c} and the disease \textit{d} are marked as the head and tail entity, respectively, we conduct database queries to retrieve the textual description of both \textit{c} and \textit{d}. 
The retrieved information is appended to the input data (see Figure~\ref{fig:model_text}).
We use the [SEP] token to separate the original input and additional text.
In cases where this led to a total number of tokens exceeding the maximum sequence length of the PLM, we first truncated the context information before truncating the input sentence.
Appendix \ref{sec:app:textual-information} provides an overview of the used databases and the number of entity mentions for which we could obtain a textual description per data set.

\subsubsection{Additional Entity Embeddings}
Next to the textual information, we evaluate the inclusion of entity embeddings trained via knowledge base embedding methods, as they are capable of encoding topological information of knowledge bases (KBs) into dense vectors that can be used to infer relations between entities in the KB \citep{ali2021bringing}. 
For this, we experimented with two methods, MuRE~\citep{balazevic2019multi} and RotatE~\citep{sun2018rotate}, trained on graphs representing interactions recorded in the CTD, i.e., chemical-gene\footnote{\url{http://ctdbase.org/reports/CTD_chem_gene_ixns.tsv.gz}}, chemical-disease\footnote{\url{https://ctdbase.org/reports/CTD_chemicals_diseases.tsv.gz}}, and chemical-phenotype interactions\footnote{\url{https://ctdbase.org/reports/CTD_pheno_term_ixns.tsv.gz}}. 
We build three knowledge graphs from the data, one having all entity types and interactions included and two exclusively focused on chemical-gene and chemical-disease relations, respectively.
%
%
%
Refer to Appendix \ref{sec:app:kg-statistics} for statistics of the used knowledge graphs.

Given an input example of our model with chemical $c$ and disease $d$, we concatenate the corresponding KBE embeddings $e_c$ and $e_d$ and feed them through a two-layer Multilayer Perceptron. 
We freeze the knowledge base embeddings while fine-tuning the PLM.
The obtained embedding $h_e$ is concatenated with the sentence embedding immediately before the output layer (see Figure~\ref{fig:model_kge}).

In addition to knowledge base embeddings, we explore the inclusion of contextual information from the mentions of the entities in the entire biomedical literature. 
This contextual information might offer additional insights into their relationships with other biomedical concepts not represented in curated knowledge bases. 
For this, we leverage dense semantic entity representations given by \cite{sanger2021large}.
Integrating literature entity embeddings is conducted similarly to the approach applied to KB embeddings.

\subsubsection{Additional Structural Information}
Finally, we assess the impact of incorporating molecular structure information in chemical- and drug-related extraction scenarios.
Like textual and embedded entity information, the similarity in drug molecular structures may offer valuable signals for predicting connections with other entities. 
To capture the molecular characteristics of chemicals,  we first retrieve their SMILES representation \citep{weininger1988smiles} from DrugBank and then explore three distinct encoding methods: molecular fingerprints, MolBERT~\citep{fabian2020molecular} and ChemBERTa~\citep{chithrananda2020chemberta}.

Molecular fingerprints represent molecules using fixed-sized binary feature vectors usually based on the presence of common substructures \citep{bolton2008pubchem} or topological information \citep{bender2004molecular}.
We integrate the fingerprints similarly to the KGE embeddings (see Figure~\ref{fig:model_kge}). 
First, we lookup the pre-computed fingerprint vector $fp_c$ for a given chemical $c$, then input $fp_c$ into a two-layer perceptron and concatenate the network output with the text embedding. 

MolBERT~\citep{fabian2020molecular} and ChemBERTa~\citep{chithrananda2020chemberta} are transformer-based compound language models trained on large compound repositories, i.e., CHEMBL \citep{gaulton2012chembl} and ZINC \citep{irwin2012zinc}, which allow computing dense vector representations of chemicals.
We adapt their original implementations\footnote{MolBERT: \url{https://github.com/BenevolentAI/MolBERT};  ChemBERTa: \url{https://github.com/seyonechithrananda/bert-loves-chemistry}} and integrate them into our baseline model as follows: for a given chemical $c$ we first retrieve the respective SMILES string $SM_c$, prepend the [CLS] token to $SM_c$ and feed it into the compound model.
Next, we use the representation of the [CLS] token in the last hidden layer as encoding for the structure of $c$ and concatenate it with the language model's output to form the classification layer's input.
Figure~\ref{fig:model_str} illustrates the integration of both models.

\section{Experiments and Results}
\subsection{Evaluation Setting}
Our study comprehensively evaluates the performance of pre-trained language models for biomedical relation extraction in a unified and consistent setting for four types of relationships: chemical-disease, chemical-gene/protein, drug-drug, and gene-disease interactions.
We evaluate our approach in two steps. 
First, we perform a hyperparameter optimization of our baseline approach using three different PLMs and one fixed random seed (907).
Based on these results, we use the three best configurations for each PLM and perform two additional runs using different seeds.
We average the results of the three runs per PLM and choose the best setting based on validation set performance.

Next, we examine the inclusion of the additional entity information using the best-performing PLM as the underlying language model.
We investigate different configuration options and hyperparameter settings using a fixed seed for each additional information.
However, we keep the hyperparameters of the base model fixed.
Analogously to the baseline models, we perform two additional runs using different seeds for the best configuration found and report the average performance. 

\subsubsection{Base models}
Our base models are built on three pre-trained language models: PubMedBERT \citep{gu2021domain}, RoBERTa-large-PM-M3-Voc \citep{lewis2020pretrained} and BioLink-BERT-Large \citep{yasunaga2022linkbert}. 
The latter two models are considerably larger in terms of parameters than the former, i.e., approximately 355M (RoBERTa-large) respectively 333M (BioLink-BERT-large) vs. 100M (PubMedBERT).
Refer to Appendix \ref{sec:app:transformer-models} for obtaining basic information (e.g. number of parameters and pre-training strategy) of the three PLMs.
We use binary cross entropy as loss function to train the base models.
We optimize our model using Adam \citep{KingmaB14} with a learning rate schedule in which the learning rate is linearly increased from zero to the target learning rate during the first 10\% of training steps and then linearly decayed to zero for the remaining 90\%. 
We explore the following hyperparameter options for the base models:  \\

\begin{compactitem}
    \item learning rate: \{5e-6, 3e-5, 5e-5\}
    \item maximum sequence length: \{256, 384, 512\}
    \item batch-size: \{8, 16, 32\}
    \item additional context sentences: \{0, 1\}
    \item task description prompt: \{yes, no\}
\end{compactitem}
\medskip

Altogether, 108 parameter configurations are evaluated for each PLM and data set.
We use the Huggingface transformers \citep{wolf2019huggingface} and PyTorch Lightning\footnote{\url{https://github.com/Lightning-AI/pytorch-lightning}} for implementing our model\footnote{Model code available at \url{https://github.com/mariosaenger/biore-kplm-benchmark}} and BigBio~\citep{fries2022bigbio} for accessing the data sets.

\subsubsection{Model extensions}
For assessing the impact of the model extensions, we leverage the best-performing PLM from the previous step and examine the following options depending on the type of additional information:
\medskip

\begin{compactitem}
    \item \textit{Additional textual information}:
    We test adding only the head entity description, only the tail description, or both to the input text.
    We append an empty string if the knowledge base does not provide a textual description for a given entity mention. 
    Since the augmentation of the input text increases its length, we also investigate higher maximum lengths, e.g., if the best baseline hyperparameter configuration has a maximum length of 384; we test 384 and 512 as maximal input lengths when adding the entity descriptions. 
    \item \textit{Additional embedded information}: We use a two-layer perceptron with hidden and output layer sizes of 100 and 0.2 as dropout probability for integrating the knowledge graph and literature embeddings.
    We train the MuRE and RotatE knowledge graph embeddings using the PyKeen library \citep{ali2021pykeen}, optimizing the hyperparameters (e.g., embedding size, learning rate, batch size) on a development set.
    Concerning the literature embeddings, we leverage the pre-computed vectors\footnote{Available at \url{https://github.com/mariosaenger/bio-re-with-entity-embeddings}} and examine all available embedding sizes (i.e., 500, 1000, 1500, and 2000 dimensions).
    For both embedding types, we explore applying distinct learning rates (i.e., \{0.001,0.0001,0.0005\}) while training the network.
    \item \textit{Structural information}: We evaluate the following fingerprint methods: atom-pair descriptors, Morgan, and RDK topological fingerprints.
    Moreover, we build a combined representation by concatenating the three fingerprints. 
    For generating the fingerprints, we leverage the RDKit library\footnote{\url{https://www.rdkit.org/}}.
    For encoding the molecular structure by MolBERT and ChemBERTa, we leverage the publicly available pre-trained models of both methods\footnote{ChemBERTa: \url{https://huggingface.co/seyonec/PubChem10M_SMILES_BPE_450k} and
    MolBERT: \url{https://ndownloader.figshare.com/files/25611290}}.
    We also investigate the impact of applying distinct learning rates (i.e., \{0.001,0.0001,0.0005\}) for both transformer networks.    
\end{compactitem}
\medskip

In total, we are conducting over 100 experiments to investigate the incorporation of additional context information for each data set. 

\subsection{Results}
\label{5:sec:results}
Table~\ref{5:tab:main_results} shows our experimental results. 
Please refer to Appendix \ref{sec:app:type-results} for relation type-specific results.
We report micro-averaged F1 scores of relation extraction and standard deviation over three runs using different random seeds for each data set and configuration. 
In the following, we discuss the base model results before reviewing the impact of incorporating the additional context information.

\subsubsection*{Base Model Results}
First, BioLinkBERT-large represents the best-performing model concerning all five data sets, followed by RoBERTa-large and PubMedBERT.
However, the differences in performance between BioLinkBERT and RoBERTa are relatively small for three of the five data sets, i.e., BC5CDR, ChemProt, and DDI.
For example, for extracting chemical-protein interactions from ChemProt, BioLinkBERT reaches an F1 score of 79.92, whereas the RoBERTa model achieves 79.82.
In contrast, the performance differences between the two large models and PubMedBERT are much more pronounced (i.e., PubMedBERT only obtains an F1 score of 77.25 on ChemProt).
However, considering the number of parameters (see Appendix~\ref{sec:app:transformer-models}), these results can certainly be expected, as the two large models have three times as many parameters as the PubMedBERT model (355M/333M vs. 100M).   
The results achieved for ChemProt and DDI are on par or even slightly better for all three language models compared to those reported in the original publications of the PLMs (see Appendix~\ref{sec:app:transformer-models}).
We attribute the performance differences to a) a broader hyperparameter search during model optimization and b) technical details regarding the representation of the input instances.

In addition to the differences between the models, we also see considerable performance variations depending on the granularity of the annotated relations, i.e., mention- or document-level.
For the former, i.e., ChemProt, CPI, and DDI, the models generally reach F1 scores of 80.0 and above, where the scores for the document-level data sets, i.e., BC5CDR and ChemDisGene, do not exceed 68.1. 
For instance, instead of considering the complete document context when making a decision, we rely on a window of two sentences.  
However, our results are nonetheless competitive, e.g., for BC5CDR, we reach an F1 score of 68.1, while state-of-the-art approaches designed explicitly for document-level extraction achieve scores of 69.4 \citep{zhou2021document}.
%
\begin{table*}[htbp]
  \centering
    \begin{tabular}{l|cc|cc|cc|cc|cc}
    \toprule
     &
      \multicolumn{2}{c|}{\textbf{BC5CDR}} &
      \multicolumn{2}{c|}{\textbf{ChemDisGene}} &
      \multicolumn{2}{c|}{\textbf{ChemProt}} &
      \multicolumn{2}{c|}{\textbf{CPI}} &
      \multicolumn{2}{c}{\textbf{DDI}}
      \\
    &
      \multicolumn{2}{c|}{(Ch-Di)} &
      \multicolumn{2}{c|}{(Ge-Di)} &
      \multicolumn{2}{c|}{(Ch-Ge)} &
      \multicolumn{2}{c|}{(Ch-Ge)} &
      \multicolumn{2}{c}{(Ch-Ch)} \\
     &
      \textbf{F1} &
      \textbf{$\sigma$} &
      \textbf{F1} &
      \textbf{$\sigma$} &
      \textbf{F1} &
      \textbf{$\sigma$} &
      \textbf{F1} &
      \textbf{$\sigma$} &
      \textbf{F1} &
      \textbf{$\sigma$}
      \\
    \midrule
    Baselines &
       &
       &
       &
       &
       &
       &
       &
       &
       &
      
      \\
      \quad PubMedBERT &
      64.16 &
      0.51 &
      52.83 &
      3.05 &
      77.25 &
      0.15 &
      81.62 &
      0.48 &
      82.10 &
      1.20
      \\
      \quad RoBERTa-Large-PM-M3-Voc &
      67.85 &
      0.70 &
      52.80 &
      2.82 &
      79.82 &
      0.60 &
      80.42 &
      0.42 &
      83.11 &
      0.19
      \\
      \quad BioLinkBERT-large &
      \textbf{68.10} &
      0.88 &
      \textbf{57.27} &
      2.72 &
      \textbf{79.92} &
      0.06 &
      \textbf{81.79} &
      1.28 &
      \textbf{83.24} &
      0.18
      \\
    \midrule
    Text &
       &
       &
       &
       &
       &
       &
       &
       &
       &
      
      \\
      \quad Chemical description &
      \cellcolor[rgb]{ 1,  .922,  .518}68.01 &
      0.13 &
      - &
      - &
      \cellcolor[rgb]{ 1,  .553,  .239}79.47 &
      0.71 &
      \cellcolor[rgb]{ 1,  .443,  .157}79.28 &
      2.91 &
      \cellcolor[rgb]{ 1,  .443,  .157}82.10 &
      0.67
      \\
      \quad Disease description &
      \cellcolor[rgb]{ 1,  .663,  .322}67.32 &
      0.36 &
      \cellcolor[rgb]{ 1,  .784,  .416}54.47 &
      1.56 &
      - &
      - &
      - &
      - &
      - &
      -
      \\
     \quad Gene description &
      - &
      - &
      \cellcolor[rgb]{ 1,  .596,  .275}52.45 &
      3.91 &
      \cellcolor[rgb]{ .941,  .91,  .49}79.98 &
      0.74 &
      \cellcolor[rgb]{ .8,  .875,  .424}82.32 &
      0.29 &
      - &
      -
      \\
      \quad Combined &
      \cellcolor[rgb]{ 1,  .733,  .376}67.51 &
      0.78 &
      \cellcolor[rgb]{ 1,  .443,  .157}50.77 &
      0.85 &
      \cellcolor[rgb]{ .761,  .863,  .404}80.15 &
      0.53 &
      \cellcolor[rgb]{ 1,  .871,  .478}81.53 &
      0.01 &
      - &
      -
      \\
    \midrule
    Entity Embeddings &
       &
       &
       &
       &
       &
       &
       &
       &
       &
      
      \\
       \quad MuRE embeddings &
      \cellcolor[rgb]{ 1,  .663,  .322}67.32 &
      0.81 &
      \cellcolor[rgb]{ 1,  .675,  .329}53.27 &
      2.16 &
      \cellcolor[rgb]{ 1,  .894,  .498}79.89 &
      0.33 &
      \cellcolor[rgb]{ 1,  .906,  .506}81.72 &
      1.06 &
      \cellcolor[rgb]{ 1,  .69,  .341}82.69 &
      0.59
      \\
       \quad RotatE embeddings &
      \cellcolor[rgb]{ 1,  .765,  .4}67.59 &
      0.51 &
      \cellcolor[rgb]{ 1,  .922,  .518}55.90 &
      0.65 &
      \cellcolor[rgb]{ .573,  .816,  .314}80.33 &
      0.56 &
      \cellcolor[rgb]{ .953,  .91,  .498}81.92 &
      1.03 &
      \cellcolor[rgb]{ 1,  .824,  .443}83.01 &
      0.35
      \\
      \quad Literature embeddings &
      \cellcolor[rgb]{ 1,  .675,  .329}67.35 &
      0.42 &
      \cellcolor[rgb]{ 1,  .675,  .329}53.26 &
      1.80 &
      \cellcolor[rgb]{ .69,  .847,  .369}80.22 &
      0.80 &
      \cellcolor[rgb]{ .788,  .871,  .416}82.36 &
      0.73 &
      \cellcolor[rgb]{ .965,  .914,  .502}83.30 &
      0.08
      \\
    \midrule
    Molecular Structure &
       &
       &
       &
       &
       &
       &
       &
       &
       &
      
      \\
       \quad Fingerprints &
      \cellcolor[rgb]{ 1,  .792,  .42}67.67 &
      0.24 &
      - &
      - &
      \cellcolor[rgb]{ 1,  .871,  .478}79.86 &
      0.51 &
      \cellcolor[rgb]{ .616,  .827,  .337}82.81 &
      0.20 &
      \cellcolor[rgb]{ .573,  .816,  .314}83.94 &
      0.55
      \\
       \quad ChemBERTa &
      \cellcolor[rgb]{ 1,  .443,  .157}66.72 &
      1.31 &
      - &
      - &
      \cellcolor[rgb]{ 1,  .443,  .157}79.33 &
      0.80 &
      \cellcolor[rgb]{ 1,  .796,  .424}81.15 &
      0.64 &
      \cellcolor[rgb]{ 1,  .898,  .498}83.19 &
      1.37
      \\
       \quad MolBERT &
      \cellcolor[rgb]{ 1,  .788,  .42}67.66 &
      0.46 &
      - &
      - &
      \cellcolor[rgb]{ 1,  .596,  .271}79.52 &
      0.54 &
      \cellcolor[rgb]{ .573,  .816,  .314}82.92 &
      0.66 &
      \cellcolor[rgb]{ 1,  .729,  .373}82.79 &
      0.95
      \\
    \bottomrule
    \end{tabular}%
  \caption{Results of our experimental evaluation. 
    For each data set we highlight the included relation scenario using (Ch) for chemicals, (Di) for diseases, and (Ge) for gene in parenthesis.
    We perform three runs for each setting and report mean F1 and standard deviation.
    First, we perform hyperparameter tuning of the three PLM base models and select the best-performing PLM per data set for evaluating the model extensions. 
    For the model extensions, we perform color-coding for each column, i.e., data set, highlighting performance improvements (compared to the best base result) in green and declines in red.
    The BioLinkBERT-large model performs best across all data sets and relation scenarios.
  }
  \label{5:tab:main_results}%
\end{table*}%

Appendix \ref{sec:app:best-hyperparameters} shows the best-performing hyperparameter setting for each PLM and data set.
Note that the best-performing settings show stark variations, both concerning an individual and multiple PLMs. 
No configuration performs best for at least two data sets, even for a single PLM.
Regarding batch size, no PLM achieves the best performance with the lowest batch size evaluated (i.e., 8). 
Similarly, the best extraction quality is reached with maximum sequence lengths of 256 and 384.
Only in one case, RoBERTa-Large with BC5CDR, does a batch size of 512 show the best results.
In two-thirds of the cases, including additional context sentences is part of the best configuration, highlighting the positive effect of extending the input text across different PLMs.
In contrast, prepending task prompts only benefit the two larger PLMs, BioLinkBERT-large and RoBERTa-large.

\subsubsection*{Model Extensions}
The investigated extensions provide only slight result improvements for only a few cases. In most settings, i.e., 30 out of 40 tested configurations, the additional information harms the base model's performance across all scenarios.
Concerning the different types of information, the results show the following impacts:\\

\begin{compactitem}
    \item \textit{Textual information}: Augmenting the input text with additional entity descriptions does not improve results in 76\% of the cases. 
        When using chemical and gene descriptions in ChemProt, the F1 score slightly increases from 79.92 to 80.15.
        However, in this setup, the variance of the model also increases substantially (SD of 0.53 vs 0.06), which calls into question the robustness of the improvements achieved.
        For the other data sets, no benefits can be recorded. 
        For ChemDisGene and DDI, considerable performance declines are recognized, e.g., from 83.24 to 82.10 (- 1.4\%) when using chemical descriptions in the case of the DDI corpus and from 57.27 to 50.77 (- 7.9\%) when using gene and disease definitions in ChemDisGene.
    \item \textit{Additional entity embeddings}: Augmenting the base model with embedded information provides a similar picture.
        Utilizing literature embeddings shows minor performance gains for DDI (83.30 vs. 83.24), CPI (81.79 vs. 82.36), and ChemProt (80.22 vs. 79.92).
        For ChemProt, including knowledge base embeddings learned via RotatE also reaches the best result (80.33 vs. 79.92).
        Considering the models' variance, however, these gains are likely attributable to statistical fluctuations.
        For BC5CDR and ChemDisGene, no performance gains can be recognized.
        Especially for the latter, sharp performance drops are recognized, e.g., augmenting the model with literature embeddings reaches an F1 score of 53.36, representing a decline by 4~pp (- 7\%) compared to the base model.
        
    \item \textit{Structural information}: The extension of the model by integrating molecular structure information positively impacts relation identification in CPI and DDI.
    Concerning DDI, the model using molecular fingerprints performs best, reaching an F1 score of 83.94, representing an improvement of 0.7~pp compared to the base model.
    Interestingly, the more advanced transformer-based language models, MolBERT and ChemBERTa, do not yield such benefits.
    For CPI, using fingerprints and MolBERT-based embeddings shows considerable performance benefits (82.81 and 82.92 vs. 81.79).  
    Again, no improvements can be recorded for the other relation extraction use cases. \\
\end{compactitem}

\section{Discussion}
\subsection{Impact of the Language Model}
Our experimental results mainly confirm the findings in \citep{weber2022chemical}, showing only minor performance enhancements using additional context information.
However, they contradict several other studies that report considerable performance improvements when integrating entity information into PLM-based models \cite{asada2021using,asada2023integrating,sousa2023k}.
One important difference between these studies and ours concerns the language models used as a base model.
Studies reporting performance improvements leverage either PubMedBERT \citep{gu2021domain}, BioBERT \cite{lee2020biobert} or SciBERT \citep{beltagy2019scibert}, In contrast, we also include larger PLMs in our evaluation.
To gain insights into the influence of the choice of language model on the results achieved when augmenting the PLMs, we carried out additional experiments using only PubMedBERT (recall that the results reported in Table \ref{5:tab:main_results} were obtained with using optimized models per setting, mostly relying on BioLinkBert-Large as base model).

Table~\ref{5:tab:pubmed_results} contains the results achieved in this setting, contrasting those obtained using the BioLinkBERT-large model.
Clear performance benefits can be recognized across all five data sets.
The augmentation of the model with entity embeddings highlights the most substantial gains.
Integrating the literature embeddings into the fine-tuning process works best, showing an average improvement of 0.9 pp in F1.
When excluding ChemDisGene, for which no improvements can be achieved, the mean score increases by an average of 1.18 pp.
Incorporating MuRE and RotatE knowledge graph embeddings represents the second and third-best configurations.

Using molecular structure information improves the extraction quality for BC5CDR and ChemProt.
For instance, leveraging the ChemBERTa model reaches an F1 score of 65.66 (+1.5pp) and 78.08 (+0.83pp) for BC5CDR and ChemProt, respectively. 
Interestingly, similar to the results using BioLinkBERT, we see no improvements for CPI, although the data set also contains chemical-protein interactions like ChemProt.
Concerning the textual context information, performance benefits can be recorded only for ChemProt.
In this case, adding gene descriptions to the input text reaches a score of 78.22, representing an increase of 0.97~pp compared to the base model.
Integrating chemical descriptions achieves slightly better results (77.7, +0.45~pp).
For the other data sets, however, textual descriptions harm the extraction performance of PubMedBERT.

To summarise, when using PubMedBERT als PLM, our study essentially supports the findings from previous studies \citep{sousa2023k,asada2021using,asada2023integrating}.
These results indicate that the larger PLMs implicitly encode (to some extent) the supervision signals obtained by the additional context information, and that the apparent contradictions in observations between our study and prior work probably can be explained by the different numbers of model parameters in the base language models. 

\subsection{Impact of Training Size}
When fine-tuning PLMs for relation extraction, a constraint is the requirement for (high-quality) training data \citep{liu2023pre}.
The annotation of relational information in biomedical texts is a complex task requiring extensive domain expertise and is, therefore, time-consuming and costly to produce \citep{huang2013comparability}.
The data sets we analyzed have between 523 (ChemDisGene) and 2432 (ChemProt) instances for training and evaluating the models (see Appendix~\ref{sec:app:data-sets}).
These considerations raise the question of how the models' performance relates to the underlying amount of training data available.
Moreover, the amount of available data can influence how the effectiveness of additional contextual information is interpreted when comparing results across studies using different datasets.
To better understand this relationship, we tested the performance of the three baseline PLM models having access only to a limited number of documents from the training set, i.e., ${25, 50, 75, ..., 200}$ training instances.
We perform six runs for each training set size and PLM using different random seeds and report mean performance of all runs.
We set the hyperparameters of the models to the best found during our original evaluation.
When building the partial training splits, we ensure a similar label distribution compared to the complete training set for data sets distinguishing multiple relation types.    

\begin{figure}[tbhp]
\includesvg[width=0.35\textwidth]{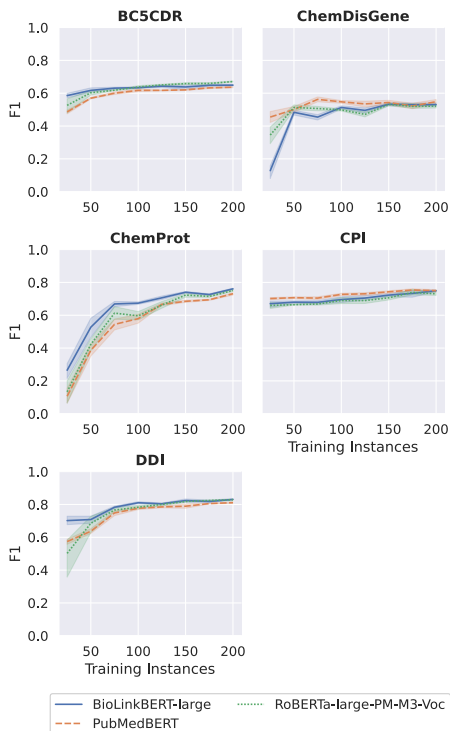}
\centering
\caption{%
Evaluation results of the models in scenarios with a reduced amount of training data.
We perform six runs for each training set size and language model using different random seeds and report mean performance of all runs.
For three of the five data sets tested, i.e., BC5CDR, CPI, and DDI, the models achieved promising results even with just 25 training documents. 
With ChemProt and ChemDisGene, the models benefit substantially from having access to more training data.
}
\label{5:fig:ablation_red}
\end{figure}

Figure~\ref{5:fig:ablation_red} illustrates the obtained results showing notable differences between the relation scenarios and data sets.
Steep increases in performance with increasing amounts of training data are particularly evident for extracting chemical-protein interactions in ChemProt and disease-gene relations in ChemDisGene.
For example, BioLinkBERT reaches an average F1 score of 27.15 when using only 25 documents from ChemProt for training, representing 34\% of the accuracy compared to the default evaluation on the entire data set.
Similarly, for ChemDisGene, the model achieves a score of 12.82 (22.4\%) when trained on 25 documents.  
However, performance improves sharply in this scenario when training on 50 documents, achieving an average score of 48.35, i.e., almost four times as high.
The performance curves of the models for the other data sets are smoother.
In the case of CPI, for instance, the models already achieve accuracies between 82\% (RoBERTa-large-PM-M3-Voc) and 85.8\% (PubMedBERT) of their full-corpus performance when trained on only 25 documents.

Concerning the ranking of the three PLM models, it is noteworthy that the results obtained during the regular evaluation (see Table \ref{5:tab:main_results}) are only consistent for three of the five data sets in the restricted evaluation setting.
For ChemDisGene and CPI, the PubMedBERT model, which has the fewest parameters of the investigated PLMs, achieves better results than the two larger models when trained on less than 150 instances. 
For example, in the case of CPI, PubMedBERT outperforms BioLinkBERT and RoBERTa-large-PM-M3-Voc by 2.7~pp and 4.3~pp, respectively, reaching an F1 score of 70.7 when restricting the training set to 50 documents.

To summarise, the sensitivity of the achieved performance to the amount of the underlying training data depends strongly on the relation scenario and the data set used. 
For three of the five data sets tested, i.e., BC5CDR, CPI, and DDI, the models achieved promising results even with just 25 training documents. 
With ChemProt and ChemDisGene, the models benefit substantially from having access to more training data.
These results indicate that differences in performance reported in prior works cannot be explained (or only to a small extent) by the size of the training data set used.
Finally, none of the performance curves in Figure~\ref{5:fig:ablation_red} converges, highlighting the potential for further performance improvements by additional training instances.

\subsection{Sentence- vs. Document-Level Relation Extraction}
\label{5:sec:mention-vs-document}
Two of the used data sets, i.e., BC5CDR and ChemDisGene, contain relation annotations given the context of the entire document (\textit{document-level}) rather than identifying relations between individual pairs of entity mentions (\textit{mention-level}).
These differences in annotation granularity usually lead to different choices when designing a model \citep{zheng2023survey}.
Accordingly, methods for relation extraction are categorized into sentence-level and document-level approaches \citep{delaunay2023comprehensive,zheng2023survey}. 
Our approach is focused on sentence-based extraction, which limits its effectiveness for document-level data sets such as BC5CDR and ChemDisGene.
Both data sets provide entity annotations on the mention level, comprising MESH identifiers for each entity mention, but have relation annotations only at the document level.
There is no annotated information telling which sentence and which specific pair of mentions express the relationship between two entities.
To compensate for this situation, we build an input instance for each pair of entities mentioned within a window of two sentences when evaluating BC5CDR and ChemDisGene.
Moreover, we compute the union of all predicted relation types for mention pairs referring to the same entities (according to their MESH identifiers) as the document-level prediction.  
To quantify the number of missed relationships of our approach, we analyze the minimal sentence distance for each relation contained in the corpora. 
For each gold standard pair, we collect all mentions of the head and tail entity and compute the shortest distance between every instance of the Cartesian product of both mention sets.     
We found that 96.34\% of the ChemDisGene and 94.16\% of the BC5CDR relations are within a context of two sentences, illustrating an upper bound of our approaches' performance.

Considering the results achieved (see Table~\ref{5:tab:main_results}), the models show large performance differences between the document- and sentence-level data sets.
For instance, the best BioLink-BERT-large base model reaches a mean F1 score of 62.69 for the two document-level and 81.65 for the three sentence-level data sets. 
To gain deeper insights into the predictions for the document-level data sets, we examined how the model's recall varies with the minimum sentence distance between the mentions of the two entities forming a gold-standard pair.
For BC5CDR, the model achieves a recall of 0.84 for intra-sentence pairs. 
In contrast, this value decreases to 0.52 for inter-sentence relations, which consist of entity pairs spanning two or more sentences.
Similarly, in the case of ChemDisGene, the model reaches 0.67 for intra- and 0.42 for inter-sentence relations. 
These findings suggest that the models encounter difficulties when dealing with more extended textual context and potential inter-sentence dependencies, highlighting the challenge of effectively capturing long-range relations.

\begin{table*}[htbp]
  \centering
    \begin{tabular}{l|cc|cc|cc|cc|cc}
    \toprule
     &
      \multicolumn{2}{c|}{\textbf{BC5CDR}} &
      \multicolumn{2}{c|}{\textbf{ChemDisGene}} &
      \multicolumn{2}{c|}{\textbf{ChemProt}} &
      \multicolumn{2}{c|}{\textbf{CPI}} &
      \multicolumn{2}{c}{\textbf{DDI}}
      \\
     &
      \multicolumn{2}{c|}{(Ch-Di)} &
      \multicolumn{2}{c|}{(Ge-Di)} &
      \multicolumn{2}{c|}{(Ch-Ge)} &
      \multicolumn{2}{c|}{(Ch-Ge)} &
      \multicolumn{2}{c}{(Ch-Ch)}
      \\
\cmidrule{2-11}     &
      \textbf{F1} &
      \textbf{$\sigma$} &
      \textbf{F1} &
      \textbf{$\sigma$} &
      \textbf{F1} &
      \textbf{$\sigma$} &
      \textbf{F1} &
      \textbf{$\sigma$} &
      \textbf{F1} &
      \textbf{$\sigma$}
      \\
        \midrule
      PubMedBERT &
      64.16 &
      0.51 &
      52.83 &
      3.05 &
      77.25 &
      0.15 &
      81.62 &
      0.48 &
      82.10 &
      1.20
      \\
    \midrule
    Textual Information &
       &
       &
       &
       &
       &
       &
       &
       &
       &
      
      \\
      \quad  Chemical Description &
      \cellcolor[rgb]{ 1,  .725,  .369}63.30 &
      0.48 &
      - &
      - &
      \cellcolor[rgb]{ .89,  .898,  .467}77.70 &
      0.47 &
      \cellcolor[rgb]{ 1,  .863,  .475}81.57 &
      0.26 &
      \cellcolor[rgb]{ 1,  .443,  .157}81.45 &
      0.63
      \\
      \quad Disease Description &
      \cellcolor[rgb]{ .949,  .91,  .494}64.34 &
      0.12 &
      \cellcolor[rgb]{ 1,  .471,  .176}48.91 &
      0.47 &
      - &
      - &
      - &
      - &
      - &
      -
      \\
      \quad Gene Description &
      - &
      - &
      \cellcolor[rgb]{ 1,  .639,  .302}50.38 &
      1.74 &
      \cellcolor[rgb]{ .573,  .816,  .314}78.22 &
      0.34 &
      \cellcolor[rgb]{ .996,  .922,  .518}81.64 &
      0.39 &
      - &
      -
      \\
      \quad Combined &
      \cellcolor[rgb]{ 1,  .443,  .157}62.03 &
      0.73 &
      \cellcolor[rgb]{ 1,  .651,  .314}50.49 &
      1.10 &
      \cellcolor[rgb]{ .953,  .91,  .498}77.60 &
      0.31 &
      \cellcolor[rgb]{ 1,  .475,  .18}81.23 &
      0.34 &
      - &
      -
      \\
        \midrule
    Entity embeddings &
       &
       &
       &
       &
       &
       &
       &
       &
       &
      
      \\
      \quad  MuRE Embeddings &
      \cellcolor[rgb]{ .678,  .843,  .365}65.30 &
      0.84 &
      \cellcolor[rgb]{ .573,  .816,  .314}53.88 &
      2.30 &
      \cellcolor[rgb]{ 1,  .922,  .518}77.52 &
      0.47 &
      \cellcolor[rgb]{ .804,  .875,  .424}82.21 &
      0.47 &
      \cellcolor[rgb]{ .737,  .859,  .392}82.83 &
      0.34
      \\
      \quad  RotatE Embeddings &
      \cellcolor[rgb]{ .663,  .839,  .357}65.35 &
      0.52 &
      \cellcolor[rgb]{ 1,  .443,  .157}48.66 &
      2.64 &
      \cellcolor[rgb]{ .812,  .878,  .427}77.83 &
      0.09 &
      \cellcolor[rgb]{ .722,  .855,  .388}82.45 &
      0.33 &
      \cellcolor[rgb]{ .929,  .906,  .486}82.30 &
      0.39
      \\
      \quad  Literature Embeddings &
      \cellcolor[rgb]{ .576,  .82,  .318}65.65 &
      0.45 &
      \cellcolor[rgb]{ 1,  .89,  .494}52.59 &
      0.48 &
      \cellcolor[rgb]{ .686,  .843,  .369}78.04 &
      0.39 &
      \cellcolor[rgb]{ .573,  .816,  .314}82.89 &
      0.21 &
      \cellcolor[rgb]{ .573,  .816,  .314}83.28 &
      0.30
      \\
    \midrule
    Molecular Structure &
       &
       &
       &
       &
       &
       &
       &
       &
       &
      
      \\
      \quad  Fingerprints &
      \cellcolor[rgb]{ .667,  .839,  .361}65.34 &
      0.42 &
      - &
      - &
      \cellcolor[rgb]{ .765,  .863,  .408}77.91 &
      0.61 &
      \cellcolor[rgb]{ .914,  .902,  .478}81.88 &
      0.78 &
      \cellcolor[rgb]{ 1,  .682,  .337}81.78 &
      0.79
      \\
      \quad  ChemBERTa &
      \cellcolor[rgb]{ .573,  .816,  .314}65.66 &
      0.27 &
      - &
      - &
      \cellcolor[rgb]{ .659,  .839,  .357}78.08 &
      0.66 &
      \cellcolor[rgb]{ 1,  .443,  .157}81.20 &
      0.90 &
      \cellcolor[rgb]{ 1,  .808,  .431}81.95 &
      0.98
      \\
      \quad  MolBERT &
      \cellcolor[rgb]{ .631,  .831,  .341}65.46 &
      0.49 &
      - &
      - &
      \cellcolor[rgb]{ .659,  .839,  .357}78.08 &
      0.40 &
      \cellcolor[rgb]{ 1,  .565,  .251}81.31 &
      0.24 &
      \cellcolor[rgb]{ 1,  .663,  .322}81.75 &
      0.76
      \\
    \bottomrule
    \end{tabular}%
  \caption{Results of our experimental evaluation using PubMedBERT as underlying baseline model. 
    For each data set we highlight the included relation scenario using (Ch) for chemicals, (Di) for diseases, and (Ge) for gene in parenthesis.
    We perform three runs for each setting and report mean F1 and standard deviation.
    For the model extensions, we perform color-coding for each column, i.e., data set, highlighting performance improvements (compared to the baseline result) in green and declines in red.
    For each data set we highlight the included relation scenario using (Ch) for chemicals, (Di) for diseases, and (Ge) for gene in parenthesis.
  }
  \label{5:tab:pubmed_results}%
\end{table*}%

\section{Conclusion}
In our study, we evaluated transformer-based language models augmented with additional context information across various biomedical relation scenarios, using five distinct datasets within a unified framework, a benchmark desperately missing so far. 
We conducted thorough hyper-parameter tuning before enhancing the best-performing models with additional entity and molecular information. 
Our results show that the BioLinkBERT-large model achieved state-of-the-art performance in multiple scenarios, though the overall benefits of including extra context were minimal. 
An ablation study further demonstrated that models with fewer parameters significantly benefit from this additional information during fine-tuning.

\section{Competing interests}
No competing interest is declared.

\section{Author contributions statement}
M.S. and U.L. conceived the experiments,  M.S. conducted the experiment, analysed the results and wrote the manuscript. M.S. and U.L. 
reviewed the manuscript.


\bibliographystyle{plain}
\bibliography{reference}

\begin{thebibliography}{10}

\bibitem{aldahdooh2024mining}
Jehad Aldahdooh, Ziaurrehman Tanoli, and Jing Tang.
\newblock Mining drug--target interactions from biomedical literature using
  chemical and gene descriptions-based ensemble transformer model.
\newblock {\em Bioinformatics advances}, 4(1):vbae106, 2024.

\bibitem{ali2021bringing}
Mehdi Ali, Max Berrendorf, Charles~Tapley Hoyt, Laurent Vermue, Mikhail Galkin,
  Sahand Sharifzadeh, Asja Fischer, Volker Tresp, and Jens Lehmann.
\newblock Bringing light into the dark: A large-scale evaluation of knowledge
  graph embedding models under a unified framework.
\newblock {\em IEEE Transactions on Pattern Analysis and Machine Intelligence},
  44(12):8825--8845, 2022.

\bibitem{ali2021pykeen}
Mehdi Ali, Max Berrendorf, Charles~Tapley Hoyt, Laurent Vermue, Sahand
  Sharifzadeh, Volker Tresp, and Jens Lehmann.
\newblock Pykeen 1.0: a python library for training and evaluating knowledge
  graph embeddings.
\newblock {\em Journal of Machine Learning Research}, 22(82):1--6, 2021.

\bibitem{asada2021representing}
Masaki Asada, Nallappan Gunasekaran, Makoto Miwa, and Yutaka Sasaki.
\newblock Representing a heterogeneous pharmaceutical knowledge-graph with
  textual information.
\newblock {\em Frontiers in Research Metrics and Analytics}, 6:670206, 2021.

\bibitem{asada2021using}
Masaki Asada, Makoto Miwa, and Yutaka Sasaki.
\newblock Using drug descriptions and molecular structures for drug--drug
  interaction extraction from literature.
\newblock {\em Bioinformatics}, 37(12):1739--1746, 2021.

\bibitem{asada2023integrating}
Masaki Asada, Makoto Miwa, and Yutaka Sasaki.
\newblock Integrating heterogeneous knowledge graphs into drug--drug
  interaction extraction from the literature.
\newblock {\em Bioinformatics}, 39(1):btac754, 2023.

\bibitem{balazevic2019multi}
Ivana Balazevic, Carl Allen, and Timothy Hospedales.
\newblock Multi-relational poincar{\'e} graph embeddings.
\newblock {\em Advances in Neural Information Processing Systems}, 32, 2019.

\bibitem{beltagy2019scibert}
Iz~Beltagy, Kyle Lo, and Arman Cohan.
\newblock Scibert: {A} pretrained language model for scientific text.
\newblock In Kentaro Inui, Jing Jiang, Vincent Ng, and Xiaojun Wan, editors,
  {\em Proceedings of the 2019 Conference on Empirical Methods in Natural
  Language Processing and the 9th International Joint Conference on Natural
  Language Processing, {EMNLP-IJCNLP} 2019}, pages 3613--3618. Association for
  Computational Linguistics, 2019.

\bibitem{bender2004molecular}
Andreas Bender, Hamse~Y Mussa, Robert~C Glen, and Stephan Reiling.
\newblock Molecular similarity searching using atom environments,
  information-based feature selection, and a naive bayesian classifier.
\newblock {\em Journal of chemical information and computer sciences},
  44(1):170--178, 2004.

\bibitem{bolton2008pubchem}
Evan~E Bolton, Yanli Wang, Paul~A Thiessen, and Stephen~H Bryant.
\newblock Pubchem: integrated platform of small molecules and biological
  activities.
\newblock In {\em Annual reports in computational chemistry}, volume~4, pages
  217--241. Elsevier, 2008.

\bibitem{brown2015gene}
Garth~R Brown, Vichet Hem, Kenneth~S Katz, Michael Ovetsky, Craig Wallin, Olga
  Ermolaeva, Igor Tolstoy, Tatiana Tatusova, Kim~D Pruitt, Donna~R Maglott,
  et~al.
\newblock Gene: a gene-centered information resource at ncbi.
\newblock {\em Nucleic acids research}, 43(D1):D36--D42, 2015.

\bibitem{chithrananda2020chemberta}
Seyone Chithrananda, Gabriel Grand, and Bharath Ramsundar.
\newblock Chemberta: large-scale self-supervised pretraining for molecular
  property prediction.
\newblock {\em arXiv preprint arXiv:2010.09885}, 2020.

\bibitem{ComparativeToxDavis2023}
Allan~Peter Davis, Thomas~C Wiegers, Robin~J Johnson, Daniela Sciaky, Jolene
  Wiegers, and Carolyn~J Mattingly.
\newblock Comparative toxicogenomics database (ctd): update 2023.
\newblock {\em Nucleic Acids Research}, 51:D1257--D1262, 1 2023.

\bibitem{delaunay2023comprehensive}
Julien Delaunay, Thi Hong~Hanh Tran, Carlos-Emiliano Gonz{\'a}lez-Gallardo,
  Georgeta Bordea, Nicolas Sidere, and Antoine Doucet.
\newblock A comprehensive survey of document-level relation extraction
  (2016-2022).
\newblock {\em arXiv preprint arXiv:2309.16396}, 2023.

\bibitem{doring2020automated}
Kersten D{\"o}ring, Ammar Qaseem, Michael Becer, Jianyu Li, Pankaj Mishra,
  Mingjie Gao, Pascal Kirchner, Florian Sauter, Kiran~K Telukunta,
  Aur{\'e}lien~FA Moumbock, et~al.
\newblock Automated recognition of functional compound-protein relationships in
  literature.
\newblock {\em Plos one}, 15(3):e0220925, 2020.

\bibitem{dou2023ik}
Mingliang Dou, Jiaqi Ding, Genlang Chen, Junwen Duan, Fei Guo, and Jijun Tang.
\newblock Ik-ddi: a novel framework based on instance position embedding and
  key external text for ddi extraction.
\newblock {\em Briefings in Bioinformatics}, 24(3):bbad099, 2023.

\bibitem{fabian2020molecular}
Benedek Fabian, Thomas Edlich, H{\'e}l{\'e}na Gaspar, Marwin Segler, Joshua
  Meyers, Marco Fiscato, and Mohamed Ahmed.
\newblock Molecular representation learning with language models and
  domain-relevant auxiliary tasks.
\newblock {\em arXiv preprint arXiv:2011.13230}, 2020.

\bibitem{fries2022bigbio}
Jason Fries, Leon Weber, Natasha Seelam, Gabriel Altay, Debajyoti Datta,
  Samuele Garda, Sunny Kang, Rosaline Su, Wojciech Kusa, Samuel Cahyawijaya,
  et~al.
\newblock Bigbio: a framework for data-centric biomedical natural language
  processing.
\newblock {\em Advances in Neural Information Processing Systems},
  35:25792--25806, 2022.

\bibitem{gaulton2012chembl}
Anna Gaulton, Louisa~J Bellis, A~Patricia Bento, Jon Chambers, Mark Davies,
  Anne Hersey, Yvonne Light, Shaun McGlinchey, David Michalovich, Bissan
  Al-Lazikani, et~al.
\newblock Chembl: a large-scale bioactivity database for drug discovery.
\newblock {\em Nucleic acids research}, 40(D1):D1100--D1107, 2012.

\bibitem{gu2021domain}
Yu~Gu, Robert Tinn, Hao Cheng, Michael Lucas, Naoto Usuyama, Xiaodong Liu,
  Tristan Naumann, Jianfeng Gao, and Hoifung Poon.
\newblock Domain-specific language model pretraining for biomedical natural
  language processing.
\newblock {\em ACM Transactions on Computing for Healthcare (HEALTH)},
  3(1):1--23, 2021.

\bibitem{hamosh2005online}
Ada Hamosh, Alan~F Scott, Joanna~S Amberger, Carol~A Bocchini, and Victor~A
  McKusick.
\newblock Online mendelian inheritance in man (omim), a knowledgebase of human
  genes and genetic disorders.
\newblock {\em Nucleic acids research}, 33:D514--D517, 2005.

\bibitem{harpaz2014text}
Rave Harpaz, Alison Callahan, Suzanne Tamang, Yen Low, David Odgers, Sam
  Finlayson, Kenneth Jung, Paea LePendu, and Nigam~H Shah.
\newblock Text mining for adverse drug events: the promise, challenges, and
  state of the art.
\newblock {\em Drug safety}, 37:777--790, 2014.

\bibitem{herrero2013ddi}
Mar{\'\i}a Herrero-Zazo, Isabel Segura-Bedmar, Paloma Mart{\'\i}nez, and
  Thierry Declerck.
\newblock The ddi corpus: An annotated corpus with pharmacological substances
  and drug--drug interactions.
\newblock {\em Journal of biomedical informatics}, 46(5):914--920, 2013.

\bibitem{huang2013comparability}
Yunda Huang and Raphael Gottardo.
\newblock Comparability and reproducibility of biomedical data.
\newblock {\em Briefings in bioinformatics}, 14(4):391--401, 2013.

\bibitem{irwin2012zinc}
John~J Irwin, Teague Sterling, Michael~M Mysinger, Erin~S Bolstad, and Ryan~G
  Coleman.
\newblock Zinc: a free tool to discover chemistry for biology.
\newblock {\em Journal of chemical information and modeling}, 52(7):1757--1768,
  2012.

\bibitem{kim2019pubchem}
Sunghwan Kim, Jie Chen, Tiejun Cheng, Asta Gindulyte, Jia He, Siqian He,
  Qingliang Li, Benjamin~A Shoemaker, Paul~A Thiessen, Bo~Yu, et~al.
\newblock Pubchem 2019 update: improved access to chemical data.
\newblock {\em Nucleic acids research}, 47(D1):D1102--D1109, 2019.

\bibitem{KingmaB14}
Diederik~P. Kingma and Jimmy Ba.
\newblock Adam: {A} method for stochastic optimization.
\newblock In Yoshua Bengio and Yann LeCun, editors, {\em 3rd International
  Conference on Learning Representations ({ICLR})}, 2015.

\bibitem{krallinger2017overview}
Martin Krallinger, Obdulia Rabal, Saber~A Akhondi, Mart{\i}n~P{\'e}rez
  P{\'e}rez, Jes{\'u}s Santamar{\'\i}a, Gael~P{\'e}rez Rodr{\'\i}guez, Georgios
  Tsatsaronis, Ander Intxaurrondo, Jos{\'e}~Antonio L{\'o}pez, Umesh Nandal,
  et~al.
\newblock Overview of the biocreative vi chemical-protein interaction track.
\newblock In {\em Proceedings of the sixth BioCreative challenge evaluation
  workshop}, volume~1, pages 141--146, 2017.

\bibitem{lai2023biorex}
Po-Ting Lai, Chih-Hsuan Wei, Ling Luo, Qingyu Chen, and Zhiyong Lu.
\newblock Biorex: Improving biomedical relation extraction by leveraging
  heterogeneous datasets.
\newblock {\em Journal of Biomedical Informatics}, 146:104487, 2023.

\bibitem{lee2020biobert}
Jinhyuk Lee, Wonjin Yoon, Sungdong Kim, Donghyeon Kim, Sunkyu Kim, Chan~Ho So,
  and Jaewoo Kang.
\newblock Biobert: a pre-trained biomedical language representation model for
  biomedical text mining.
\newblock {\em Bioinformatics}, 36(4):1234--1240, 2020.

\bibitem{lewis2020pretrained}
Patrick Lewis, Myle Ott, Jingfei Du, and Veselin Stoyanov.
\newblock Pretrained language models for biomedical and clinical tasks:
  understanding and extending the state-of-the-art.
\newblock In {\em Proceedings of the 3rd Clinical Natural Language Processing
  Workshop}, pages 146--157, 2020.

\bibitem{li2016biocreative}
Jiao Li, Yueping Sun, Robin~J Johnson, Daniela Sciaky, Chih-Hsuan Wei, Robert
  Leaman, Allan~Peter Davis, Carolyn~J Mattingly, Thomas~C Wiegers, and Zhiyong
  Lu.
\newblock Biocreative v cdr task corpus: a resource for chemical disease
  relation extraction.
\newblock {\em Database}, 2016, 2016.

\bibitem{lipscomb2000medical}
Carolyn~E Lipscomb.
\newblock Medical subject headings (mesh).
\newblock {\em Bulletin of the Medical Library Association}, 88(3):265, 2000.

\bibitem{liu2023pre}
Pengfei Liu, Weizhe Yuan, Jinlan Fu, Zhengbao Jiang, Hiroaki Hayashi, and
  Graham Neubig.
\newblock Pre-train, prompt, and predict: A systematic survey of prompting
  methods in natural language processing.
\newblock {\em ACM Computing Surveys}, 55(9):1--35, 2023.

\bibitem{luo2024learning}
Yizhen Luo, Kai Yang, Massimo Hong, Xing~Yi Liu, Zikun Nie, Hao Zhou, and
  Zaiqing Nie.
\newblock Learning multi-view molecular representations with structured and
  unstructured knowledge.
\newblock In {\em Proceedings of the 30th ACM SIGKDD Conference on Knowledge
  Discovery and Data Mining}, pages 2082--2093, 2024.

\bibitem{maglott2010entrez}
Donna Maglott, Jim Ostell, Kim~D Pruitt, and Tatiana Tatusova.
\newblock Entrez gene: gene-centered information at ncbi.
\newblock {\em Nucleic acids research}, 39(suppl\_1):D52--D57, 2010.

\bibitem{mcinnes2024biobert}
Bridget~T McInnes, Jiawei Tang, Darshini Mahendran, and Mai~H Nguyen.
\newblock Biobert-based deep learning and merged chemprot-drugprot for enhanced
  biomedical relation extraction.
\newblock {\em arXiv preprint arXiv:2405.18605}, 2024.

\bibitem{sanger2021large}
Mario S{\"a}nger and Ulf Leser.
\newblock Large-scale entity representation learning for biomedical
  relationship extraction.
\newblock {\em Bioinformatics}, 37(2):236--242, 2021.

\bibitem{sousa2023k}
Diana~F Sousa and Francisco~M Couto.
\newblock K-ret: knowledgeable biomedical relation extraction system.
\newblock {\em Bioinformatics}, 39(4):btad174, 2023.

\bibitem{sun2018rotate}
Zhiqing Sun, Zhi-Hong Deng, Jian-Yun Nie, and Jian Tang.
\newblock Rotate: Knowledge graph embedding by relational rotation in complex
  space.
\newblock In {\em International Conference on Learning Representations}, 2018.

\bibitem{tang2024mollm}
Xiangru Tang, Andrew Tran, Jeffrey Tan, and Mark~B Gerstein.
\newblock Mollm: a unified language model for integrating biomedical text with
  2d and 3d molecular representations.
\newblock {\em Bioinformatics}, 40(Supplement\_1):i357--i368, 2024.

\bibitem{uniprot2023uniprot}
{UniProt Consortium and others}.
\newblock Uniprot: the universal protein knowledgebase in 2021.
\newblock {\em Nucleic acids research}, 49(D1):D480--D489, 2021.

\bibitem{weber2022chemical}
Leon Weber, Mario S{\"a}nger, Samuele Garda, Fabio Barth, Christoph Alt, and
  Ulf Leser.
\newblock Chemical--protein relation extraction with ensembles of carefully
  tuned pretrained language models.
\newblock {\em Database}, 2022, 2022.

\bibitem{wei2024pubtator}
Chih-Hsuan Wei, Alexis Allot, Po-Ting Lai, Robert Leaman, Shubo Tian, Ling Luo,
  Qiao Jin, Zhizheng Wang, Qingyu Chen, and Zhiyong Lu.
\newblock Pubtator 3.0: an ai-powered literature resource for unlocking
  biomedical knowledge.
\newblock {\em Nucleic Acids Research}, page gkae235, 2024.

\bibitem{wei2019pubtator}
Chih-Hsuan Wei, Alexis Allot, Robert Leaman, and Zhiyong Lu.
\newblock Pubtator central: automated concept annotation for biomedical full
  text articles.
\newblock {\em Nucleic acids research}, 47(W1):W587--W593, 2019.

\bibitem{weininger1988smiles}
David Weininger.
\newblock Smiles, a chemical language and information system. 1. introduction
  to methodology and encoding rules.
\newblock {\em Journal of chemical information and computer sciences},
  28(1):31--36, 1988.

\bibitem{wolf2019huggingface}
Thomas Wolf, Lysandre Debut, Victor Sanh, Julien Chaumond, Clement Delangue,
  Anthony Moi, Pierric Cistac, Tim Rault, R{\'e}mi Louf, Morgan Funtowicz,
  et~al.
\newblock Huggingface's transformers: State-of-the-art natural language
  processing.
\newblock {\em arXiv preprint arXiv:1910.03771}, 2019.

\bibitem{yasunaga2022linkbert}
Michihiro Yasunaga, Jure Leskovec, and Percy Liang.
\newblock Linkbert: Pretraining language models with document links.
\newblock In {\em Proceedings of the 60th Annual Meeting of the Association for
  Computational Linguistics (Volume 1: Long Papers)}, pages 8003--8016.
  Association for Computational Linguistics, 2022.

\bibitem{zhang-etal:2022:LREC}
Dongxu Zhang, Sunil Mohan, Michaela Torkar, and Andrew McCallum.
\newblock A distant supervision corpus for extracting biomedical relationships
  between chemicals, diseases and genes.
\newblock In {\em Proceedings of The 13th Language Resources and Evaluation
  Conference}. European Language Resources Association, 2022.

\bibitem{zheng2023survey}
Yifan Zheng, Yikai Guo, Zhizhao Luo, Zengwen Yu, Kunlong Wang, Hong Zhang, and
  Hua Zhao.
\newblock A survey on document-level relation extraction: Methods and
  applications.
\newblock In {\em 3rd International Conference on Internet, Education and
  Information Technology (IEIT 2023)}, pages 1061--1071. Atlantis Press, 2023.

\bibitem{zhou2014biomedical}
Deyu Zhou, Dayou Zhong, and Yulan He.
\newblock Biomedical relation extraction: From binary to complex.
\newblock {\em Computational and Mathematical Methods in Medicine}, 2014:1--18,
  2014.

\bibitem{zhou2021document}
Wenxuan Zhou, Kevin Huang, Tengyu Ma, and Jing Huang.
\newblock Document-level relation extraction with adaptive thresholding and
  localized context pooling.
\newblock In {\em Proceedings of the AAAI conference on artificial
  intelligence}, volume~35, pages 14612--14620, 2021.

\end{thebibliography}

\clearpage

\begin{appendices}
\renewcommand{\thefigure}{A.\arabic{figure}}
\setcounter{figure}{0} 

\setcounter{table}{0}
\renewcommand{\thetable}{AT.\arabic{table}}

\onecolumn

\section{Appendix}
\subsection{A.1: Data Sets}
\label{sec:app:data-sets}

\textbf{BC5CDR:} For chemical-disease relationships, we leverage the BioCreative V CDR corpus (BC5CDR) \citep{li2016biocreative} containing 1,500 PubMed articles with 15,953 annotated chemicals, 13,318 diseases, and 3,169 chemical-disease interactions.
The data set provides entity identifiers for both entity types using Medical Subject Headings (MeSH) on mention level.
In contrast, relation annotations are given only on document-level, i.e., instead of identifying relations between individual pairs of entity mentions (\textit{mention-level}), the corpus provides chemical-disease interactions within the context of entire documents. 
Moreover, the data set does not discern different subtypes of chemical-disease relations. \\

\textbf{ChemProt:} We use the ChemProt corpus \citep{krallinger2017overview}, created in the context of the BioCreative Challenge VI in 2016.
The data set consists of 2,482 scientific abstracts from PubMed and identifies 32,514 chemical and 30,922 protein mentions and their interactions, differentiated into nine different types.
Within the evaluation of the challenge, these interaction types were divided into five groups (see Table~\ref{tab:5_chemprot_reltypes}).
In our approach, we follow the evaluation protocol of the challenge by using the five interaction groups as gold standard annotations. \\

\textbf{CPI}: As a second data set for chemical-protein interactions, we utilize the relations given in the compound-protein corpus (CPI)~\citep{doring2020automated}.
The data set consists of 2,613 sentences from 1,808 PubMed abstracts containing 3,563 chemical and 4,256 protein mentions. 
In total, 2,931 chemical-protein interactions are annotated.
In contrast to the ChemProt data set, no distinct types of chemical-protein interactions are distinguished. \\

\textbf{ChemDisGene}: For evaluating the extraction of gene-disease relationships, we use the curated corpus provided by the ChemDisGene data set \citep{zhang-etal:2022:LREC} containing 523 scientific abstracts from PubMed. 
The data set comprises 5,739 chemicals, 2,931 diseases, and 5,578 gene annotations and interactions between them.
Chemicals and diseases are normalized to CTD chemicals and CTD diseases using MESH identifiers.
Genes are identified using NCBI Gene.
Our study uses the 576 document-level gene-disease associations from the data set, further categorized into \textit{marker/mechansim} and \textit{therapeutic} relations.
\\

\textbf{DDI corpus}: For the identification of drug-drug interactions, we utilize the DDI corpus \citep{herrero2013ddi}, which contains 158 MedLine abstracts and 683 texts from the DrugBank database describing over 5,000 interactions between 17,783 drug mentions.
The data set distinguishes four types of drug-drug interactions: \textit{mechanism}, \textit{effect}, \textit{advice}, and \textit{interaction}, whereby the last type represents a general relationship between two drugs without a more detailed description. \\

\subsection{A.2: Entity Normalization Statistics}
\label{sec:app:entity-normalization}
%
\begin{table}[H]
  \centering
    \begin{tabular}{ll|cc|cc}
    \toprule
     &
       &
      \multicolumn{2}{c|}{\textbf{Entitiy Mentions}} &
      \multicolumn{2}{c}{\textbf{Unique Entities}}
      \\
    \multicolumn{1}{c}{\textbf{Data Set}} &
      \multicolumn{1}{c|}{\textbf{Entity Type}} &
      \textbf{Normalized} &
      \textbf{Ratio} &
      \textbf{Normalized} &
      \textbf{Ratio}
      \\
    \midrule
    ChemProt &
      Chemical &
      25.566 &
      78.63\% &
      5.297 &
      69.35\%
      \\
     &
      Gene/Protein &
      22.268 &
      72.01\% &
      5.415 &
      56.84\%
      \\
      \midrule
    CPI &
      Chemical &
      3.306 &
      92.79\% &
      1.326 &
      91.45\%
      \\
     &
      Gene/Protein &
      3.092 &
      72.65\% &
      1.162 &
      70.90\%
      \\
        \midrule
    DDI &
      Chemical &
      15.698 &
      88.28\% &
      2.528 &
      73.57\%
      \\
    \bottomrule
    \end{tabular}
   \caption{Overview of the annotations of BC5CDR, CPI and DDI that we could map to shared ontologies, i.e.,  NCBI Gene \citep{brown2015gene} for genes and CTD Chemicals \citep{ComparativeToxDavis2023} for chemicals.
   For each data set and entity type we report the number of normalized mentions and unique entities (identified by distinct strings) as well as their ratio with regard to the total number of entities in the data set.
   }
  \label{5:tab:entity_normalization}%
\end{table}%

\clearpage

\subsection{A.3: Relation labels}
\label{sec:app:relation-labels}
\begin{table}[H]
    \centering
    \begin{tabular}{llp{3cm}}
            \toprule
         \textbf{Relation} &
         \textbf{Relation Label} &
         \textbf{Interaction Types} \\
            \midrule
        CPR:3 &
        upregulator &
        upregulator, activator, indirect upregulator \\         
        
        CPR:4 &
        downregulator &
        downregulator, inhibitor, indirect downregulator \\
        
        CPR:5 &
        agonist &
        agonist, agonist-activator, agonist-inhibitor \\

        CPR:6 & 	
        antagonist &
        antagonist \\
        
        CPR:9 &
        substrate &
        substrate, product of \\
        
            \bottomrule
    \end{tabular}
    \caption{Overview about the relation and interaction types of the ChemProt corpus \citep{krallinger2017overview} with chemical-protein relationships.}
    \label{tab:5_chemprot_reltypes}
\end{table}

\subsection{A.4: Textual Information}
\label{sec:app:textual-information}
\begin{table}[htbp]
  \centering
    \begin{tabular}{r|c|r|r}
    \toprule
    \multicolumn{1}{c|}{\textbf{Data Set}} &
      \textbf{Chemicals} &
      \multicolumn{1}{c|}{\textbf{Diseases}} &
      \multicolumn{1}{c}{\textbf{Genes}}
      \\
    \midrule
    \multicolumn{1}{l|}{BC5CDR} &
      15,210 &
      \multicolumn{1}{c|}{13,192} &
      
      \\
     &
      \textit{(95.34\%)} &
      \multicolumn{1}{c|}{\textit{(99.05\%)}} &
      
      \\
    \midrule
    \multicolumn{1}{l|}{ChemDisGene} &
       &
      \multicolumn{1}{c|}{2,836} &
      \multicolumn{1}{c}{5,437}
      \\
     &
       &
      \multicolumn{1}{c|}{(96.76\%)} &
      \multicolumn{1}{c}{(97.48\%)}
      \\
      \midrule
    \multicolumn{1}{l|}{ChemProt} &
      24,892 &
       &
      \multicolumn{1}{c}{20,574}
      \\
     &
      \textit{(78,20\%)} &
       &
      \multicolumn{1}{c}{\textit{(67,87\%)}}
      \\
    \midrule
    \multicolumn{1}{l|}{CPI} &
      3.306 &
       &
      \multicolumn{1}{c}{3.092}
      \\
     &
      (92,79\%) &
       &
      \multicolumn{1}{c}{(72.65\%)}
      \\
    \midrule
    \multicolumn{1}{l|}{DDI corpus} &
      16,301 &
       &
      
      \\
     &
      \textit{(88.16\%)} &
       &
      
      \\
    \bottomrule
    \end{tabular}%
    \caption{Number of entity mentions per data set for which additional textual descriptions could be found in the respective knowledge base.
    Note that the knowledge bases do not necessarily contain a textual description for each entity they list. 
    Hence, we can not extract textual descriptions for each entity even if the data set provides entity identifiers (e.g., BC5CDR and ChemDisGene).
    }
    \label{5:tab:text-data}%
\end{table}%

\subsection{A.5: Knowledge Graph Statistics}
\label{sec:app:kg-statistics}
\begin{table}[H]
  \centering
    \begin{tabular}{l|r|r|r}
    \toprule
     &
      \multicolumn{3}{c}{\textbf{Knowledge Graph}}
      \\
     &
      \multicolumn{1}{c|}{\textbf{Complete}} &
      \multicolumn{1}{c|}{\textbf{Chemical-Disease}} &
      \multicolumn{1}{c}{\textbf{Chemical-Gene}}
      \\
    \midrule
    Triples &
      1,131,375 &
      1,033,060 &
      78,272
      \\
        \midrule
    Entities &
       &
       &
      
      \\
       \quad Chemicals &
      16,265 &
      15,752 &
      7,646
      \\
       \quad Diseases &
      7,068 &
      7,071 &
      -
      \\
       \quad Genes &
      15,097 &
      - &
      15,176
      \\
       \quad Phenotypes &
      2,797 &
      - &
      -
      \\
       \quad Total &
      41,227 &
      22,823 &
      22,822
      \\
        \midrule
    Relation Types &
      124 &
      1 &
      115
      \\
    \bottomrule
    \end{tabular}%
  \caption{Statistics of the CTD-based knowledge graphs that we use for training knowlege base embeddings using MuRE~\citep{balazevic2019multi} and RotatE~\citep{sun2018rotate}.
  Note, the high number of relation types for the complete and the chemical-gene graph results from the fine-granular modelling of relationships in CTD, e.g., for many interactions also the effect is recorded in the relation type (e.g., \textit{increases-expression} and \textit{decreases-expression}).    
  }
  \label{5:tab:kge_data}%
\end{table}%

\clearpage

\subsection{A.6: Transformer Models}
\label{sec:app:transformer-models}
\begin{table}[H]
  \centering
    \begin{tabular}{l|c|c|c}
    \toprule
     &
      \textbf{PubMedBERT} &
      \textbf{RoBERTa-Large-} &
      \textbf{BioLinkBERT}
      \\
     &
       &
      \textbf{PM-M3-Voc} &
      \textbf{-large}
      \\
     &
      (\citep{gu2021domain}) &
      (\citep{lewis2020pretrained}) &
      (\citep{yasunaga2022linkbert})
      \\
    \midrule
    Parameters &
      $\sim$ 100M &
      $\sim$ 355M &
      $\sim$ 333M
      \\
        \midrule
    Pre-training &
       &
       &
      
      \\
       \quad  Data  &
      PubMed &
      PubMed+PMC+&
      PubMed
      \\
      &&
      MIMIC-III
      \\
       \quad  Method &
      from scratch &
      from scratch &
      from scratch
      \\
       \quad  Tasks &
      MLM+NSP &
      MLM+NSP &
      MLM+NSP+DRP
      \\
        \midrule
    RE Results &
       &
       &
      
      \\
      \quad ChemProt &
      77.24 &
      76.2 &
      79.98
      \\
      \quad DDI &
      82.36 &
      82.1 &
      83.35
      \\
    \bottomrule
    \end{tabular}%
    \caption{Overview of the evaluated language models. 
       For each model, we report the number of parameters and pre-training information.
       Moreover, we include the relation extraction results reported in the original article of each PLM for data sets used in this study.
       Pre-training tasks include masked language modeling (MLM), next sentence prediction (NSP), and document relation prediction (DRP).
       All models are pre-trained from scratch instead of continuing training of existing PLMs.
    }
    \label{5:tab:transformers}%
\end{table}%

\subsection{A.7: Best-performing hyperparameters}
\label{sec:app:best-hyperparameters}
\begin{sidewaystable}[hp]
  \centering
    \begin{tabular}{lccccc}
    \toprule
    \textbf{PLM / Data Set} &
      \multicolumn{1}{l}{\textbf{Learning Rate}} &
      \multicolumn{1}{l}{\textbf{Batch Size}} &
      \multicolumn{1}{l}{\textbf{Max. Sequence Length}} &
      \multicolumn{1}{l}{\textbf{Context Size}} &
      \multicolumn{1}{l}{\textbf{Task Prompt}}
      \\
    \midrule
    PubMedBERT &
       &
       &
       &
       &
      
      \\
    \; BC5CDR &
      0.00003 &
      32 &
      256 &
      1 &
      -
      \\
    \; ChemDisGene &
      0.00005 &
      32 &
      384 &
      0 &
      -
      \\
    \; ChemProt &
      0.00005 &
      32 &
      384 &
      1 &
      -
      \\
    \; CPI &
      0.00005 &
      16 &
      256 &
      0 &
      -
      \\
    \; DDI &
      0.00005 &
      32 &
      256 &
      1 &
      -
      \\
    \midrule
    RoBERTa-Large-PM-M3-Voc &
       &
       &
       &
       &
      
      \\
    \; BC5CDR &
      0.000005 &
      32 &
      512 &
      1 &
      \checkmark
      \\
    \; ChemDisGene &
      0.00003 &
      32 &
      384 &
      1 &
      -
      \\
    \; ChemProt &
      0.00003 &
      32 &
      256 &
      1 &
      -
      \\
    \; CPI &
      0.00003 &
      16 &
      256 &
      0 &
      \checkmark
      \\
    \; DDI &
      0.00003 &
      32 &
      384 &
      1 &
      -
      \\
    \midrule
    BioLinkBERT-large &
       &
       &
       &
       &
      
      \\
    \; BC5CDR &
      0.00003 &
      16 &
      384 &
      1 &
      \checkmark
      \\
    \; ChemDisGene &
      0.000005 &
      16 &
      384 &
      1 &
      -
      \\
    \; ChemProt &
      0.00005 &
      32 &
      256 &
      1 &
      \checkmark
      \\
    \; CPI &
      0.00005 &
      16 &
      256 &
      0 &
      \checkmark
      \\
    \; DDI &
      0.00003 &
      32 &
      384 &
      0 &
      -
      \\
    \bottomrule
    \end{tabular}%
  \caption{Best-performing hyperparameter setting per PLM and data set.}
  \label{5:tab:hyperparameters}%
\end{sidewaystable}%

\clearpage

\subsection{A.8 Fine-grained Relation Type Results}
\label{sec:app:type-results}
The gold-standard annotations of the ChemProt and DDI corpus distinguish between five types of chemical-protein interactions and four types of drug-drug interactions, respectively.
The results in Section~\ref{5:sec:results} only show an aggregated picture of the performance of the different PLMs and the extensions investigated, not accounting for relation type-specific performance differences in these two scenarios.
We also investigated the type-specific scores to gain a more fine-grained understanding of how well the PLMs and the proposed model extensions perform.
For this experiment, we used the best-performing PLM for each scenario, i.e., BioLinkBERT-large, and the best-performing configuration for each type of model extension, i.e., textual side information, embedded side information, and structure information.
We excluded the ChemDisGene data set, which distinguishes two types of gene-disease interactions due to its small size.

Table~\ref{5:tab:reltype_results} highlights the relation type-specific results that vary considerably between individual relation types, e.g., from 70.88 (\textit{Substrate}) to 84.51 (\textit{Downregulator}) for ChemProt and from 58.37 (\textit{Interaction}) to 90.44 (\textit{Advise}) for DDI in case of the baseline model. 
Regarding the ChemProt results, it should first be noted that BioLinkBERT-large achieves surprisingly good results even for relation types with low support.
For \textit{Antagonist}, which represents the type with the second-fewest instances, the model achieves the second-best score (83.68).
However, the highest score is reached for the majority type \textit{Downregulator}.
When we inspect the results of the model extensions, we see a rather heterogeneous picture concerning the different types of enhancements.
Text and structure information are beneficial for the low-support relation types.
For instance, augmenting the model with structure information improves the F1 score by 0.56~pp for \textit{Agonist} and 1.57~pp for \textit{Antagonist}.
In contrast, utilizing embedded context information is advantageous for \textit{Substrate} relations, lifting the score by 1.31~pp.
These diverse results suggest that ensembling models with different extension types could lead to improvements across different relation types.

The picture of the results for DDI is different.
Performance increases can primarily be reached for \textit{Effect} relations by enhancing the model with embedded entity (81.57, +0.75~pp) and structure information (81.55, +0.73~pp).
Using additional textual descriptions provides no performance boosts for any type of interaction.
Moreover, the extraction of \textit{Mechanism} and \textit{Advice} is not improved by any additional context information.
Sometimes, even sharp performance declines are recognized, e.g., -1.5~pp for \textit{Mechanism} interactions when using textual entity descriptions.

%
\begin{table*}[htbp]
  \centering
    \begin{tabular}{rlc|cc|x{1.25cm}x{1.25cm}|x{1.25cm}x{1.25cm}|x{1.25cm}x{1.25cm}}
    \toprule
     &
       &
       &
      \multicolumn{2}{c|}{\textbf{Baseline}} &
      \multicolumn{2}{c|}{\textbf{Textual}} &
      \multicolumn{2}{c|}{\textbf{Embedded}} &
      \multicolumn{2}{c}{\textbf{Structure}}
      \\
     &
       &
       &
      \multicolumn{2}{c|}{\textbf{}} &
      \multicolumn{2}{c|}{\textbf{Information}} &
      \multicolumn{2}{c|}{\textbf{Information}} &
      \multicolumn{2}{c}{\textbf{Information}}
      \\
    \multicolumn{1}{l}{\textbf{Data Set}} &
      \multicolumn{1}{c}{\textbf{Relation Type}} &
      \textbf{Support} &
      \textbf{F1} &
      \textbf{$\sigma$} &
      \textbf{F1} &
      \textbf{$\sigma$} &
      \textbf{F1} &
      \textbf{$\sigma$} &
      \textbf{F1} &
      \textbf{$\sigma$}
      \\
    \midrule
    \multicolumn{1}{l}{ChemProt} &
      Agonist &
      195 &
      75.67 &
      1.62 &
      \cellcolor[rgb]{ .573,  .816,  .314}76.38 &
      0.85 &
      \cellcolor[rgb]{ 1,  .443,  .157}73.73 &
      1.13 &
      \cellcolor[rgb]{ .667,  .839,  .357}76.23 &
      0.04
      \\
     &
      Antagonist &
      293 &
      83.68 &
      1.46 &
      \cellcolor[rgb]{ .843,  .882,  .443}84.73 &
      0.48 &
      \cellcolor[rgb]{ 1,  .922,  .518}84.42 &
      1.33 &
      \cellcolor[rgb]{ .573,  .816,  .314}85.25 &
      0.86
      \\
     &
      Downregulator &
      1661 &
      84.51 &
      0.24 &
      \cellcolor[rgb]{ .769,  .867,  .408}84.87 &
      0.46 &
      \cellcolor[rgb]{ .573,  .816,  .314}85.17 &
      0.62 &
      \cellcolor[rgb]{ 1,  .443,  .157}84.04 &
      0.67
      \\
     &
      Upregulator &
      665 &
      76.56 &
      0.76 &
      \cellcolor[rgb]{ 1,  .922,  .518}75.89 &
      0.79 &
      \cellcolor[rgb]{ 1,  .878,  .486}75.85 &
      0.07 &
      \cellcolor[rgb]{ 1,  .443,  .157}75.41 &
      0.93
      \\
     &
      Substrate &
      644 &
      70.88 &
      1.10 &
      \cellcolor[rgb]{ 1,  .443,  .157}70.50 &
      1.29 &
      \cellcolor[rgb]{ .573,  .816,  .314}72.19 &
      0.48 &
      \cellcolor[rgb]{ .667,  .839,  .361}71.91 &
      1.32
      \\
    \midrule
    \multicolumn{1}{l}{DDI} &
      Advise &
      221 &
      90.44 &
      1.03 &
      \cellcolor[rgb]{ 1,  .443,  .157}88.95 &
      2.06 &
      \cellcolor[rgb]{ 1,  .839,  .455}89.79 &
      0.74 &
      \cellcolor[rgb]{ 1,  .922,  .518}89.96 &
      3.05
      \\
     &
      Effect &
      360 &
      80.82 &
      1.16 &
      \cellcolor[rgb]{ 1,  .443,  .157}80.13 &
      0.38 &
      \cellcolor[rgb]{ .573,  .816,  .314}81.57 &
      0.96 &
      \cellcolor[rgb]{ .584,  .82,  .322}81.55 &
      0.82
      \\
     &
      Interaction &
      96 &
      58.37 &
      0.95 &
      \cellcolor[rgb]{ 1,  .671,  .329}56.68 &
      2.70 &
      \cellcolor[rgb]{ .573,  .816,  .314}58.48 &
      1.25 &
      \cellcolor[rgb]{ 1,  .443,  .157}55.11 &
      1.50
      \\
     &
      Mechanism &
      302 &
      87.02 &
      1.57 &
      \cellcolor[rgb]{ 1,  .443,  .157}85.52 &
      1.35 &
      \cellcolor[rgb]{ 1,  .922,  .518}86.59 &
      0.70 &
      \cellcolor[rgb]{ 1,  .588,  .267}85.85 &
      0.54
      \\
    \bottomrule
    \end{tabular}%
   \caption{Relation type-specific evaluation results of the best baseline model, i.e., BioLinkBERT-large, and the best-performing model enhancement per type of context information (i.e. textual, embedded and structure information).
   For each row, we apply color-coding indicating performance increases (green) and declines (green) compared to the baseline model results.
   }
  \label{5:tab:reltype_results}%
\end{table*}%

\end{appendices}

\end{document}